\documentclass[10pt,twocolumn,letterpaper]{article}

\usepackage{wacv}
\usepackage{times}
\usepackage{epsfig}
\usepackage{graphicx}
\usepackage{amsmath}
\usepackage{amssymb}

\usepackage{subcaption, multirow, booktabs, url, flushend}
\usepackage[table]{xcolor}

\usepackage[pagebackref=true,breaklinks=true,letterpaper=true,colorlinks,bookmarks=false]{hyperref}

\newcolumntype{L}[1]{>{\raggedright\let\newline\\\arraybackslash\hspace{0pt}}m{#1}}
\newcolumntype{C}[1]{>{\centering\let\newline\\\arraybackslash\hspace{0pt}}m{#1}}
\newcolumntype{R}[1]{>{\raggedleft\let\newline\\\arraybackslash\hspace{0pt}}m{#1}}

\wacvfinalcopy 

\def\wacvPaperID{202}

\ifwacvfinal\fi
\begin{document}

\title{Performance of Humans in Iris Recognition:\\The Impact of Iris Condition and Annotation-driven Verification\thanks{Paper accepted for WACV 2019, Hawaii, USA. The authors would like to thank Dr.~Sidney D'Mello and Mr.~Robert Bixler for making their laboratory and eye tracker device available for conducting the experiments.}}

\author{
    Daniel Moreira,
    Mateusz Trokielewicz,
    Adam Czajka,
    Kevin W. Bowyer, and
    Patrick J. Flynn\thanks{
D.~Moreira, A.~Czajka, K.~Bowyer, and P.~Flynn are with the Department of Computer Science and Engineering, Univ.~of Notre Dame, USA.
\newline
\indent M.~Trokielewicz is with the Biometrics Laboratory, Research and Academic Computer Network (NASK), Poland.
\newline
\indent Corresponding author: Dr.~Daniel Moreira (dhenriq1@nd.edu).
}
}

\maketitle
\ifwacvfinal\thispagestyle{empty}\fi

\begin{abstract}
\vspace{-0.22cm}
This paper advances the state of the art in human examination of iris images by (1)~assessing the impact of different iris conditions in identity verification, and (2)~introducing an annotation step that improves the accuracy of people's decisions.
In a first experimental session, 114 subjects were asked to decide if pairs of iris images depict the same eye (genuine pairs) or two distinct eyes (impostor pairs).
The image pairs sampled six conditions: (1)~easy for algorithms to classify, (2)~difficult for algorithms to classify, (3)~large difference in pupil dilation, (4)~disease-affected eyes, (5)~identical twins, and (6)~post-mortem samples.
In a second session, 85 of the 114 subjects were asked to annotate matching and non-matching regions that supported their decisions. 
Subjects were allowed to change their initial classification as a result of the annotation process.
Results suggest that: (a)~people improve their identity verification accuracy when asked to annotate matching and non-matching regions between the pair of images, (b)~images depicting the same eye with large difference in pupil dilation were the most challenging to subjects, but benefited well from the annotation-driven classification, (c)~humans performed better than iris recognition algorithms when verifying genuine pairs of post-mortem and disease-affected eyes (i.e., samples showing deformations that go beyond the distortions of a healthy iris due to pupil dilation), 
and (d)~annotation does not improve accuracy of analyzing images from identical twins, which remain 
confusing for people.
\end{abstract}
\vspace{-0.22cm}

\section{Introduction}
\label{sec:intro}
\vspace{-0.14cm}

The literature of iris recognition has been investigating the performance of humans at tasks such as iris texture perception~\cite{stark_2010, bowyer_2010, hollingsworth_2010, hollingsworth_2011, shen_2013} and identity verification~\cite{mcginn_2013, guest_2013}.
Understanding how people perceive and analyze iris features is useful not only for inspiring the development of better solutions, but also for making them more \emph{human-intelligible}.

Human-intelligible iris recognition is particularly necessary in forensic applications, where experts often rely on the outputs of algorithms for sustaining conclusions and presenting them in a court of law.
As pointed out by Chen \etal~\cite{chen_2016}, in spite of the traditional iris recognition solutions providing nearly-perfect false match rates~\cite{daugman_2006}, they are yet far from being human-friendly enough to convince people who do not possess image processing expertise.

Moreover, human-intelligible iris recognition helps to meet the need to deploy more transparent and accountable systems~\cite{nissembaum_1994, wachter_2017}.
As stated in the new European General Data Protection Regulation (GDPR)~\cite{gdpr_2018}, citizens have the right to obtain an explanation of decisions made about them by algorithms belonging to either government or industry.
As a consequence, iris recognition solutions lacking human-intelligible decision processes may face usage hindrances. 
Indeed, the accountability discussion is also present in the American scientific community, proven by the recent efforts of the National Science Foundation (NFS) in funding research on the topic~\cite{cornell_2017}.

\begin{table*}[t]
\renewcommand{\arraystretch}{1.2}
\caption{Literature of human performance in iris recognition. The present work is added to the last row.}
\label{tab:rw}
\centering
\footnotesize
\begin{tabular}{L{1.7cm} L{1.4cm} C{1.1cm} C{1.2cm} C{1.3cm} L{2.3cm} L{5.5cm}}
\hline
\rowcolor{gray!8}{\bf Reference} & {\bf Problem} & {\bf Subjects (\#)} & {\bf Trials per session (\#)} & {\bf Average session time (min)} & {\bf Used images} & {\bf Data details} \\
\hline
Stark {\it et~al.}~\cite{stark_2010} & Iris texture perception & 21 & 100 & 30 & Segmented iris only & 100~images~depicting~100~distinct~irises~from 100 distinct individuals\\
\cmidrule(lr){1-7}
Bowyer {\it et~al.}~\cite{bowyer_2010} & Iris texture perception & 55 & 210 & 10$^\dagger$ & Whole eye, segmented iris only, or periocular only & 630~images~depicting~630~distinct~irises~from 315 distinct individuals \\
\cmidrule(lr){1-7}
Hollingsworth {\it et~al.}~\cite{hollingsworth_2010} & Iris texture perception & 28 & 196 & 10$^\dagger$ & Segmented iris only or periocular only & 392 images depicting 392 distinct irises from 196 distinct individuals (including twins' pairs)\\
\cmidrule(lr){1-7}
Shen and Flynn~\cite{shen_2013} & Iris texture perception & 21$^\ddagger$ & 64$^\ddagger$ & 270 & Strip-normalized iris & 124 images depicting 62 distinct irises from 62 distinct individuals\\
\cmidrule(lr){1-7}
McGinn {\it et~al.}~\cite{mcginn_2013} & Identity verification & 22 & 190 & 26 & Whole eye & 202 images depicting 109 distinct irises from 109 distinct individuals (including twins' pairs)\\
\cmidrule(lr){1-7}
Guest {\it et~al.}~\cite{guest_2013} & Identity verification & 32 & 52 & 10 & Whole eye & 208 images depicting 104 distinct irises from 104 distinct individuals\\
\cmidrule(lr){1-7}
\textbf{This work} & {\bf Identity verification} & {\bf 114} & {\bf 30$^\ast$, 24$^\star$} & {\bf 17$^\ast$, 15$^\star$} & {\bf Segmented iris only} & {\textbf{1360 images of 512 distinct irises from 512 individuals (with varied pupil dilation, twins', disease-affected, and post-mortem samples)}}\\
\hline
\end{tabular}\\
\vspace{0.2cm}
$\dagger$ Lower-bound estimated value; each subject had three seconds to inspect each trial --- $\ddagger$ Average value of three sessions\\
$\ast$ Conducted at the University of Notre Dame --- $\star$ Conducted at the Research and Academic Computer Network (NASK)
\vspace{-0.1cm}
\end{table*}

This paper contributes to the understanding of how an everyman performs identity verification based on iris patterns.
In each of two experiments, subjects are presented a pair of irises and asked to decide whether the pair belongs to the same eye.
In the first experiment, we apply a typical multiple-choice questionnaire~\cite{mcginn_2013, guest_2013}, with no request for image regions or features that justify the decisions.
In the second experiment, subjects are asked to manually annotate matching and non-matching regions in the pair of irises, which support their decisions, as an effort to make them more conscious about the task.

In the experiments, there are image pairs representing six different conditions, which are either commonplace or reportedly known to pose challenges to automated systems or to human examiners: (1)~healthy eyes that are easily handled by an example iris recognition software, (2)~healthy eyes that are challenging for the same software, (3)~disease-affected eyes, (4)~iris pairs with extensive difference in pupil dilation, (5)~irises of identical twins, and (6)~iris images acquired from deceased individuals.
This variety of conditions allowed us to observe that pairs of images depicting the same iris but with different pupil dilation, iris images of twins, and post-mortem samples are the most challenging to humans.
Also, subjects were able to improve their recognition accuracy when they were asked to manually annotate regions supporting their decision.
That was not true, however, in the case of iris images of identical twins, which were so confusing that the assessed numbers of improved and worsened decisions were similar.

In summary, this paper advances the state of the art in human examination of iris images with the following contributions:

\begin{itemize}
    \item Assessment of human skills in verifying the identity of iris images presenting different conditions, including healthy eyes of unrelated individuals, of identical twins, and never used before disease-affected and post-mortem iris samples.
    
    \item Employment of custom software to allow subjects to annotate the image regions they rely upon to classify an iris pair, and analysis of how this helps them to provide more accurate decisions.
    
    \item Introduction of the notation of \emph{non-matching regions}, besides the typical concept of \emph{matching regions}, in the process of matching pairs of iris images.
\end{itemize}

The remainder of this paper has four sections.
In Sec.~\ref{sec:rw}, we discuss the related work, while in Sec.~\ref{sec:es}, we detail the configuration of experiments.
In Sec.~\ref{sec:results}, in turn, we report the obtained results, followed by Sec.~\ref{sec:conclusions}, where we discuss the lessons learned from the experiments.

\section{Related Work}
\label{sec:rw}
\vspace{-0.1cm}

There are only a few works related to human examination of iris images.
Stark \etal~\cite{stark_2010} studied how people classify iris textures into categories.
They used a software tool that allowed subjects to browse a set of segmented near-infrared iris images and use a drag-and-drop scheme to organize the images into groups based on their perception of the iris textures.
They found that people consistently identify similar categories and subcategories of irises.

\begin{figure*}[t]
\centering
\includegraphics[width=17.4cm]{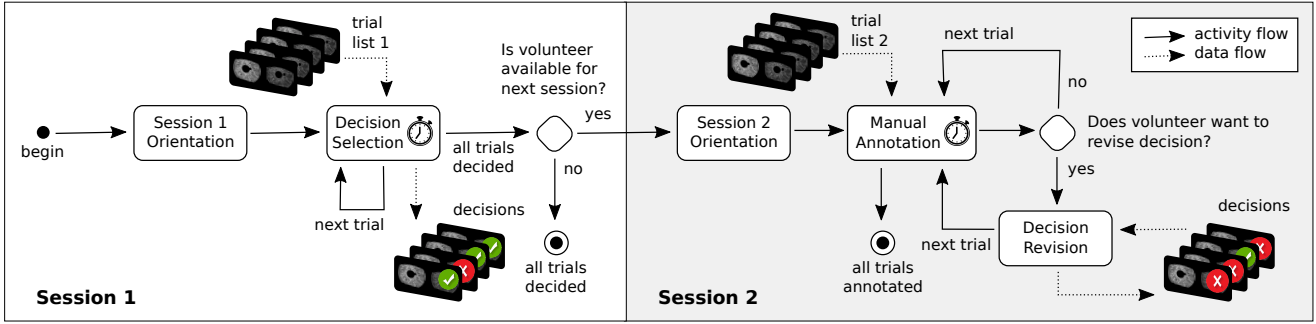}
\caption{Experimental methodology overview.
Rounded rectangular boxes denote subjects' activities, solid arrows
represent their precedence, and dashed arrows denote data flow.
Experiments always begin with \emph{Session~1}, namely the annotation-less experimental part.
Subjects available for \emph{Session~2} then participate in the annotation-driven experimental part.
}
\label{fig:pipeline}
\end{figure*}

Bowyer \etal~\cite{bowyer_2010} investigated people's ability to recognize right and left irises as belonging to the same person or not.
Through experiments, they discovered that humans perceive texture similarities that are not detected by automated solutions.
As a consequence, they can correctly guess, with only three seconds viewing, if left and right eyes belong to the same individual.
When evaluating near-infrared images of the whole eye, subjects achieved an accuracy of 86\% in the task at hand.
When evaluating images with iris portions masked out, subjects achieved an accuracy of 83\%, by relying only on the periocular parts of samples.
Subjects' ratings of image pairs were collected using a five-level scale, ranging from (1)~``same individual (certain)'', (2)~``same individual (likely)'', (3)~``uncertain'', (4)~``different people (likely)'', to (5)~``different people (certain)''.

In a similar fashion, Hollingsworth \etal~\cite{hollingsworth_2010} investigated people's skills in deciding if two different iris images depict the eyes of twin siblings or not.
Contrary to the typical identity verification pipeline~\cite{daugman_2004}, which the authors reported as being useless for the task at hand, human examiners could reach an accuracy of 81\% when spending only three seconds analyzing pairs of near-infrared segmented iris images.
The accuracy dropped to 76.5\% when only periocular regions were available.
Again, subjects' responses were collected using a five-level rating.
Hollingsworth \etal~\cite{hollingsworth_2011} present the combined findings of \cite{bowyer_2010} and \cite{hollingsworth_2010}.

Shen and Flynn~\cite{shen_2013} asked people to manually annotate \emph{iris crypts}, oval-shaped iris regions with strong edges and darker interior, over near-infrared images.
Using annotation software, subjects were asked to outline the borders of the crypts, finding ``easy'' and ``challenging'' samples, depending on the clarity of crypts.
Presented images comprised strip-normalized iris images, with non-iris-texture regions masked out.
The aim of the research was to figure out the utility of crypts for developing more human-interpretable iris-based identity verification.
For that, they assessed the repeatability of annotated crypts across subjects, finding that it was possible in the case of ``easy'' samples.

McGinn \etal~\cite{mcginn_2013} assessed the performance of human examiners in iris-based identity verification.
For that, they asked subjects to classify pairs of irises as either genuine (two images depicting the same eye) or impostor (two images depicting different eyes), again with a five-level rating scale.
Presented images comprised near-infrared samples, containing whole eyes of either close-age and same-ethnicity unrelated individuals, or of identical twins.
They concluded that identical twins pose a challenge to human performance.
In spite of that, the overall accuracy was very high: 92\% of the time subjects were successful in classifying iris pairs.
Finally, results suggested that subjects improved skills as they gained experience.

Guest \etal~\cite{guest_2013}, in turn, investigated the performance of humans in deciding if two distinct infrared whole-eye images depict the same eye or not.
In the experiments, subjects presented an overall decision accuracy of 83.2\%.

\begin{figure}[t]
\centering
\includegraphics[width=0.47\textwidth]{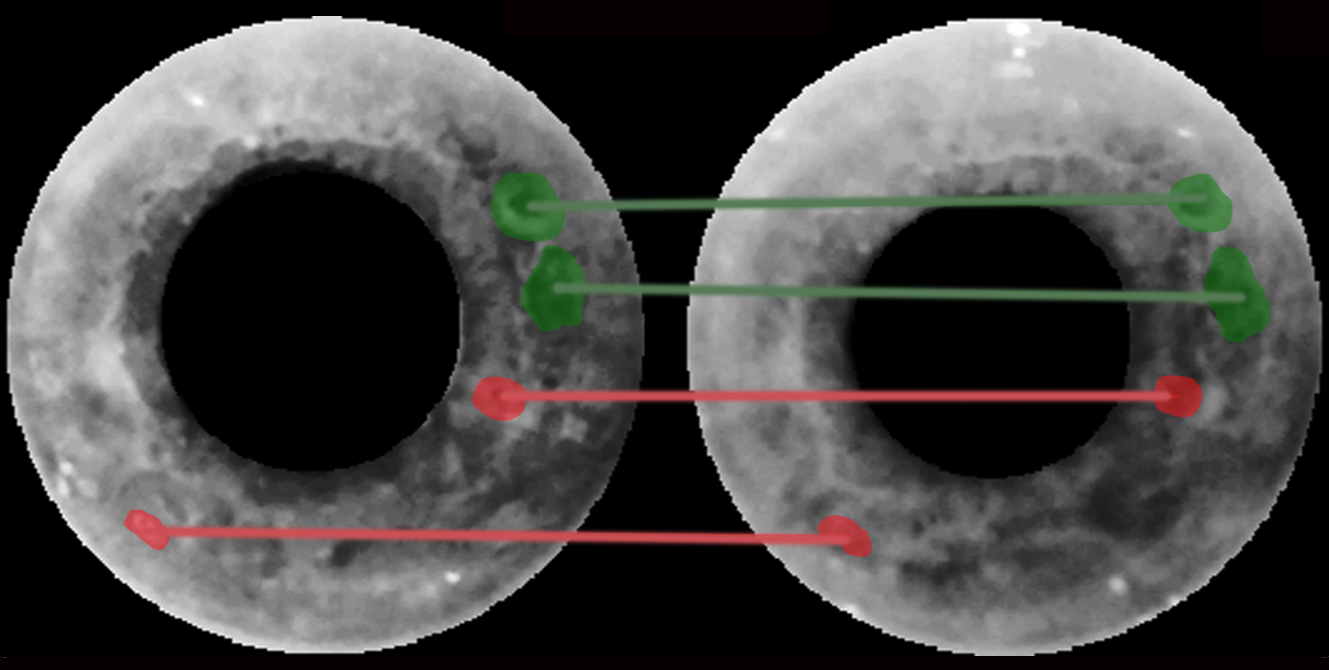}
\caption{An example of manual annotation containing matching (in green) and non-matching (in red) regions between two post-mortem irises.}
\label{fig:annotation}
\vspace{-0.2cm}
\end{figure}

Table~\ref{tab:rw} summarizes these previous works and the work described in this paper.
To our knowledge, there are no other publications about human examination of iris images.

\section{Experimental Setup}
\label{sec:es}
\vspace{-0.1cm}

\begin{figure*}[t]
    \centering
    \begin{subfigure}[t]{0.24\textwidth}
        \fcolorbox{black}{black}{\includegraphics[width=0.43\textwidth]{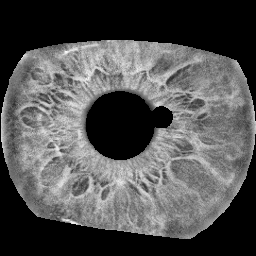}}
        \fcolorbox{black}{black}{\includegraphics[width=0.43\textwidth]{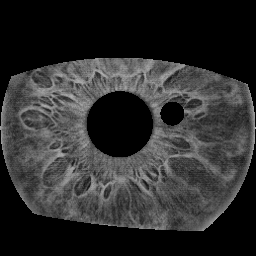}}
        \caption{}
    \end{subfigure}\hfill
    \begin{subfigure}[t]{0.24\textwidth}
        \fcolorbox{black}{black}{\includegraphics[width=0.43\textwidth]{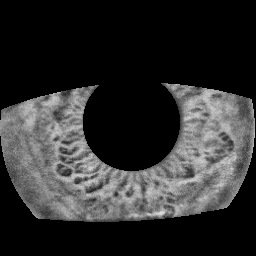}}
        \fcolorbox{black}{black}{\includegraphics[width=0.43\textwidth]{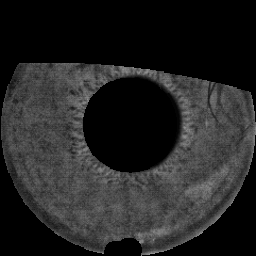}}
        \caption{}
    \end{subfigure}\hfill
    \begin{subfigure}[t]{0.24\textwidth}
        \fcolorbox{black}{black}{\includegraphics[width=0.43\textwidth]{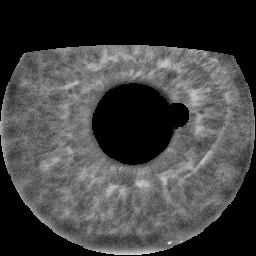}}
        \fcolorbox{black}{black}{\includegraphics[width=0.43\textwidth]{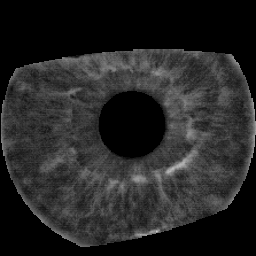}}
        \caption{}
    \end{subfigure}\hfill
    \begin{subfigure}[t]{0.24\textwidth}
        \fcolorbox{black}{black}{\includegraphics[width=0.43\textwidth]{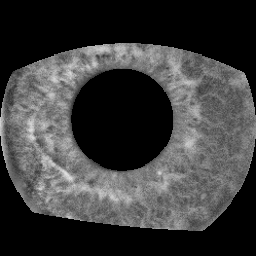}}
        \fcolorbox{black}{black}{\includegraphics[width=0.43\textwidth]{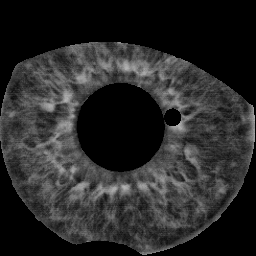}}
        \caption{}
    \end{subfigure}\vskip2mm

    \begin{subfigure}[t]{0.24\textwidth}
        \fcolorbox{black}{black}{\includegraphics[width=0.43\textwidth]{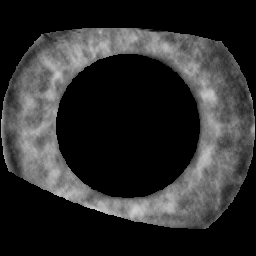}}
        \fcolorbox{black}{black}{\includegraphics[width=0.43\textwidth]{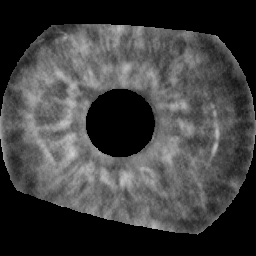}}
        \caption{}
    \end{subfigure}\hfill
    \begin{subfigure}[t]{0.24\textwidth}
        \fcolorbox{black}{black}{\includegraphics[width=0.43\textwidth]{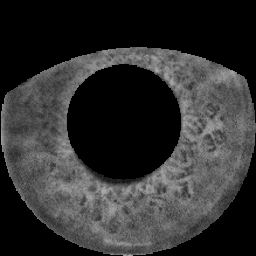}}
        \fcolorbox{black}{black}{\includegraphics[width=0.43\textwidth]{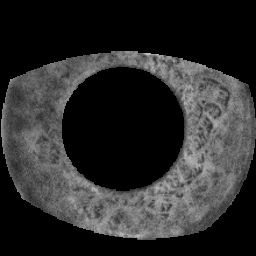}}
        \caption{}
    \end{subfigure}\hfill
    \begin{subfigure}[t]{0.24\textwidth}
        \fcolorbox{black}{black}{\includegraphics[width=0.43\textwidth]{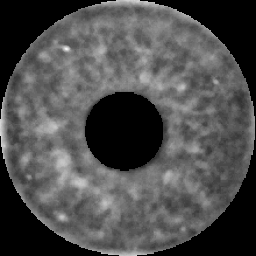}}
        \fcolorbox{black}{black}{\includegraphics[width=0.43\textwidth]{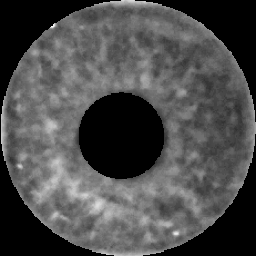}}
        \caption{}
    \end{subfigure}\hfill
    \begin{subfigure}[t]{0.24\textwidth}
        \fcolorbox{black}{black}{\includegraphics[width=0.43\textwidth]{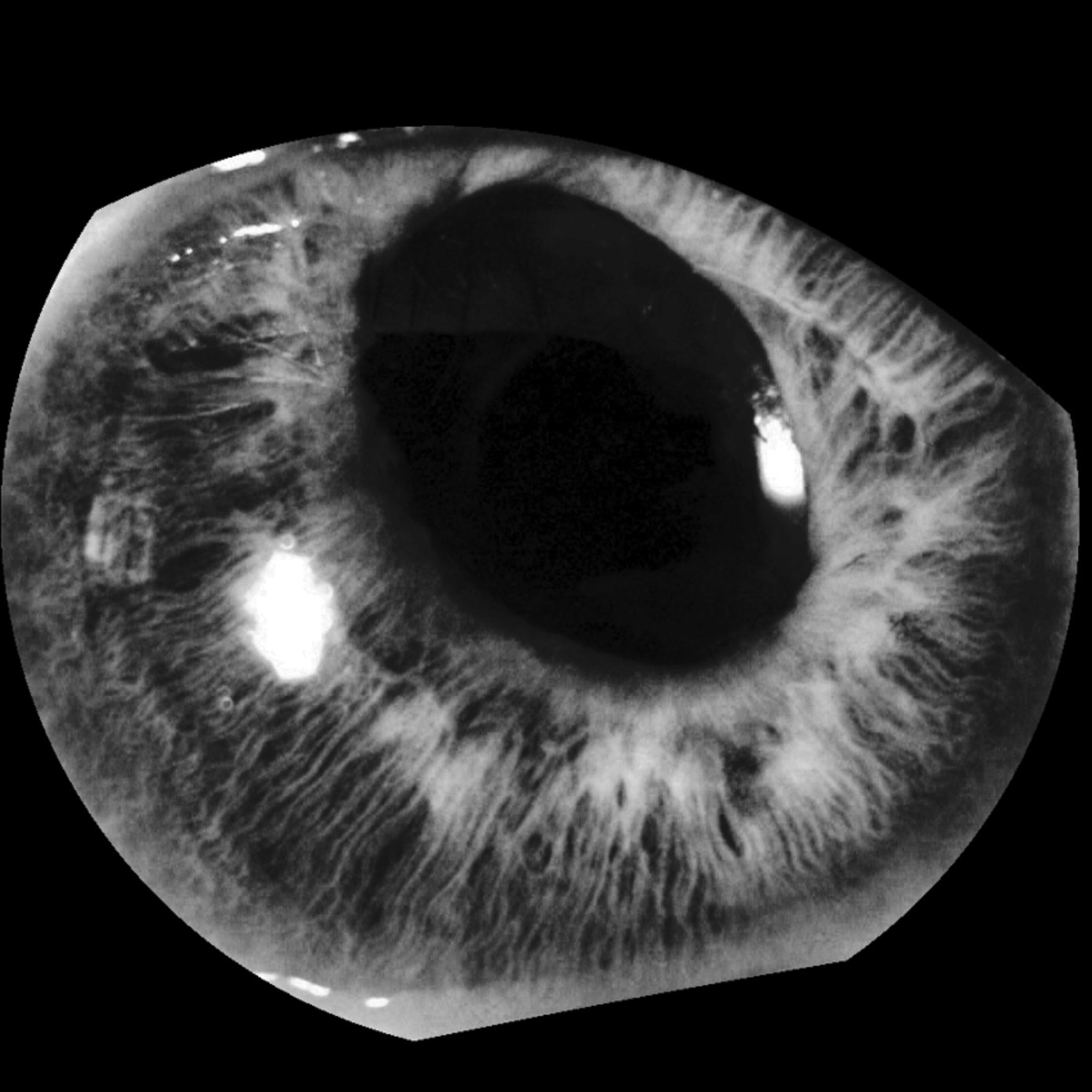}}
       \fcolorbox{black}{black}{\includegraphics[width=0.43\textwidth]{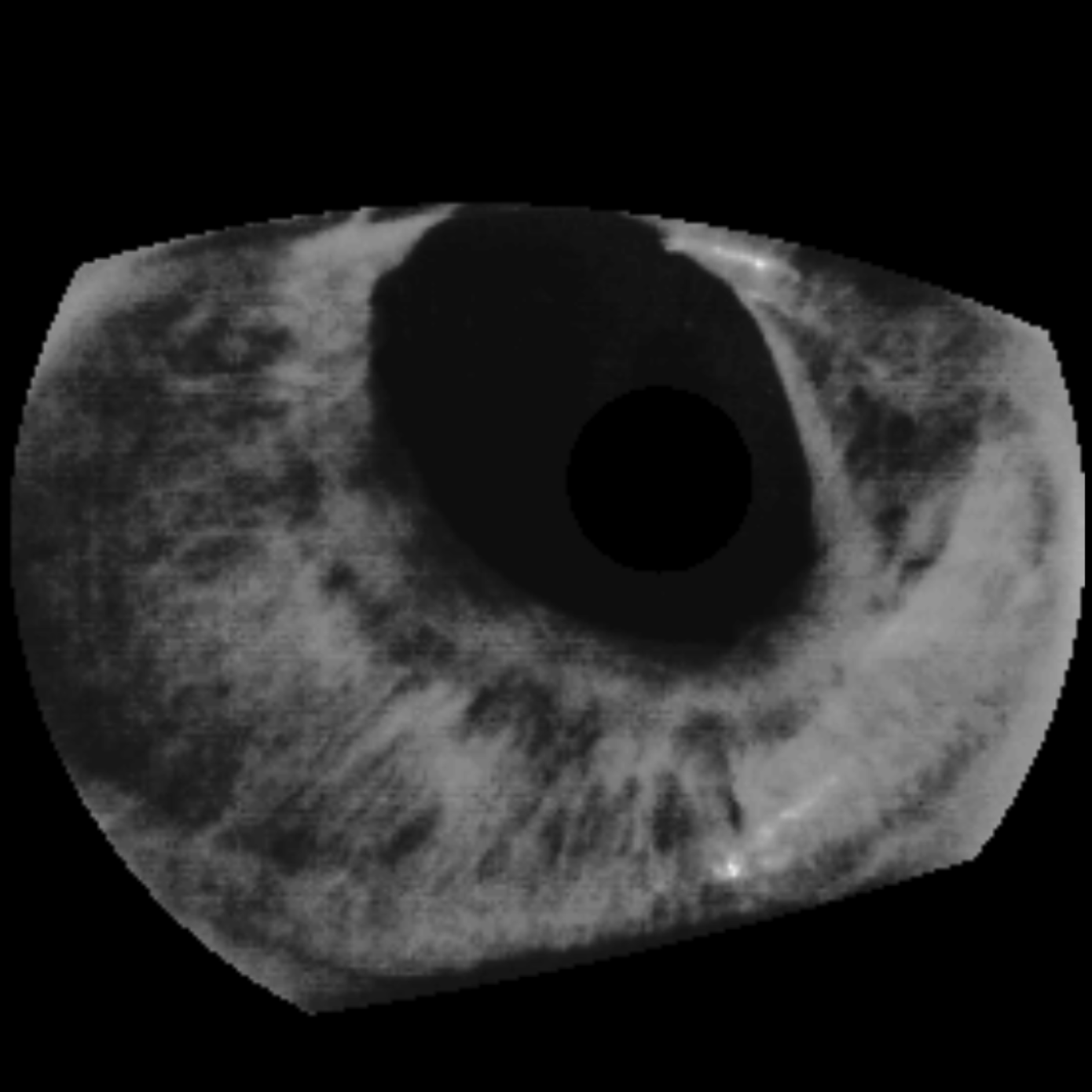}}
        \caption{}
    \end{subfigure}
    \caption{Examples of iris pairs presented to subjects: (a)~pair with the same iris generating {\it low} Hamming distance, (b)~different irises generating {\it high} Hamming distance, (c)~pair with the same iris image generating {\it high} Hamming distance, (d)~different irises generating {\it low} Hamming distance, (e)~the same iris before and after {\it visible light stimulation}, (f)~different irises of {\it identical twins}, (g)~the same {\it post-mortem} iris captured five and 16 hours after death, (h)~the same disease-affected iris captured in two different sessions.}
    \label{fig:samples:print_and_lens}
\vspace{-0.4cm}
\end{figure*} 

The experimental setup is described in five parts. 
In Sec.~\ref{sec:method}, we introduce the two-session experimental methodology, while in Sec.~\ref{sec:dataset} we explain the chosen categories of iris pairs, including data sources and image pre-processing.
In Sec.~\ref{sec:nd}~and~\ref{sec:wu}, respectively, we describe the experiments conducted at the University of Notre Dame and at NASK headquarters.
Finally, in Section~\ref{sec:os}, we detail the experimental setup for employing \emph{OSIRIS}~\cite{othman_2016}, which contains an open-source implementation of Daugman's method for iris recognition~\cite{daugman_2004}, and \emph{IriCore}~\cite{iricore_2018} and \emph{MIRLIN}~\cite{mirlin_2018},
two state-of-the-art solutions for iris recognition.
The idea is to provide, along with the performance of humans, the results of fully automated strategies.

\subsection{Experimental methodology}
\label{sec:method}
\vspace{-0.1cm}

We propose a two-session experimental method that allows humans to perform identity verification through the examination of iris patterns.
For that, we collect subjects' decisions on whether iris image pairs depict the same eye or not.
In the first session, subjects are expected only to provide their decision, with no need for clarification.
In the second session, subjects are asked to provide a manual annotation of the image regions that they see as matching or diverging between the pair of iris images, in order to justify their classification of the image pair.
This serves as an effort to make them more conscious about the task at hand.
Fig.~\ref{fig:pipeline} provides an overview of the proposed experimental method, with the activities that we envision for each subject.

\emph{Session~1} starts with an orientation and signing of the consent form (\emph{Session 1 Orientation} in Fig.~\ref{fig:pipeline}).
The subject then views a sequence of image pairs and judges whether or not they represent the same eye (\emph{Decision Selection} in Fig.~\ref{fig:pipeline}).
The list of trials (\emph{trial list 1} in Fig.~\ref{fig:pipeline}) is previously generated through a pre-processing step, in which irises are selected and genuine and impostor pairs are created.
Details about the selected samples are presented in Sec.~\ref{sec:dataset}.
Decisions are then recorded and stored for further analysis.
Subjects are given as much time as they need to make a decision, and the decision times are collected for each trial (explaining the chronometer icons in Fig.~\ref{fig:pipeline}).
This first session was followed in experiments at two different institutions, which allowed us to get more diverse results across distinct subject populations.

\emph{Session~2} is an annotation-driven part of the experiment that subjects are asked if they have time to complete.
Subjects who elect to do this session receive additional instructions on how to provide manual annotation.
A subset of iris image pairs used in the first session is selected for annotation in this second session (\emph{trial list 2} in Fig.~\ref{fig:pipeline}).
Details about the selected samples are given in Sec.~\ref{sec:dataset}.
For each trial, the subject is asked to annotate and connect matching and non-matching regions between the two irises.
Subjects are also given access to their decision made in \emph{Session~1}, and allowed to change such decision.
As in \emph{Session~1}, subjects can spend as much time as they need in this task, and the time intervals are recorded.
Fig.~\ref{fig:annotation} depicts an example of manual annotation of matching and non-matching regions between two postmortem irises.

\subsection{Dataset of iris image pairs}
\label{sec:dataset}

To assess the influence of different conditions on the accuracy of human iris recognition, we conducted experiments with six categories of irises, which are either commonplace or reportedly known to pose challenges to automated systems or human examiners:
\vspace{-0.1cm}

\begin{enumerate}
    \item \textit{\textbf{Healthy and easy:}} images depicting apparently healthy eyes that pose no challenge to the OSIRIS iris recognition software used in this study~\cite{othman_2016}.
    That is, the case of genuine pairs generating low Hamming distances between them (Fig.~\ref{fig:samples:print_and_lens}(a)), or of impostor pairs whose iris codes yield large Hamming distances (Fig.~\ref{fig:samples:print_and_lens}(b)).
    \vspace{-0.1cm}
    
    \item \textit{\textbf{Healthy but difficult:}} apparently healthy eyes that pose challenges to the OSIRIS software.
    That is, the case of genuine pairs generating unexpectedly large Hamming distances (Fig.~\ref{fig:samples:print_and_lens}(c)), or of impostor pairs generating unexpectedly small Hamming distances (Fig.~\ref{fig:samples:print_and_lens}(d)).
    \vspace{-0.1cm}
    
     \item \textit{\textbf{Large difference in pupil dilation:}} images of the same eye with significantly difference in pupil dilation, as representatives of the natural iris transformations that occur due to variations in environment lighting (Fig.~\ref{fig:samples:print_and_lens}(e)).
     \vspace{-0.1cm}
    
    \item \textit{\textbf{Twins:}} images depicting different eyes, one from each of a pair of identical twins, which are reportedly recognizable by humans, in opposition to being indifferent to automated systems~\cite{hollingsworth_2011} (Fig.~\ref{fig:samples:print_and_lens}(f)).
    \vspace{-0.1cm}
    
    \item \textit{\textbf{Post-mortem:}} images depicting either the same or different eyes, captured from deceased individuals, which are known to be surprisingly useful for iris recognition \cite{Trokielewicz_2016} (Fig.~\ref{fig:samples:print_and_lens}(g)).
    \vspace{-0.1cm}
    
     \item \textit{\textbf{Disease-affected:}} images depicting the same eye, which suffers from varied eye diseases that may deteriorate the recognition reliability of automated systems~\cite{Trokielewicz_2015} (Fig.~\ref{fig:samples:print_and_lens}(h)).
     \vspace{-0.1cm}
\end{enumerate}

\begin{figure}[t]
\centering
\includegraphics[width=0.47\textwidth]{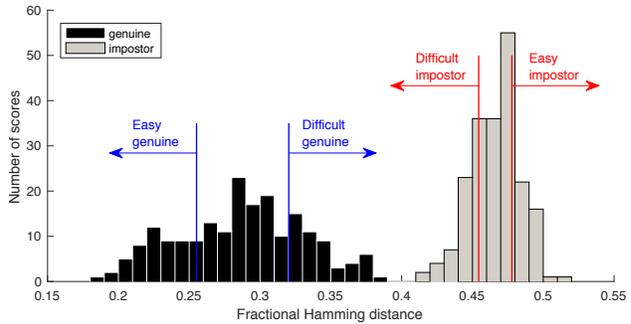}
\vspace{-0.2cm}
\caption{Distributions of the fractional Hamming distance between iris codes. 
Codes were calculated using the OSIRIS software.
These distributions were used to select ``easy'' and ``difficult'' cases to use in experiments.}
\label{fig:easy-difficult-distro}
\vspace{-0.2cm}
\end{figure}

Fig.~\ref{fig:easy-difficult-distro} illustrates the distributions of genuine and impostor comparison scores generated by OSIRIS to image pairs of healthy eyes.
This information was used to select ``easy'' and ``difficult'' cases.
Additionally, we generated both genuine and impostor pairs for disease-affected and post-mortem eyes.
With respect to twins' samples, it was obviously not possible to generate genuine pairs.
To balance the number of impostor and genuine trials, we did not generate impostor pairs from images presenting a large difference in pupil size.
Also, when generating the genuine and impostor pairs of healthy, post-mortem, and disease-affected irises, we neither mixed different categories, nor created pairs of images that were captured on the same day. 

Given that our intent was to focus on the iris texture and that the dataset was very diverse, we manually segmented all the images and masked out the regions that should not be used by subjects in their judgment, such as eyelashes, eyelids, specular reflections, and severe effects from disease or post-mortem deterioration (\eg corneal wrinkles).
In addition, contrast-limited adaptive histogram equalization (CLAHE~\cite{pizer_1987}) was used to enhance contrast for image display, as illustrated in Fig.~\ref{fig:samples:print_and_lens}.

Images of healthy eyes were collected from the \emph{ND-CrossSensor-Iris-2013} dataset~\cite{ndcsdataset_2013}.
Disease-affected iris images were picked from \emph{Warsaw-BioBase-Disease-Iris v2.1} database \cite{Trokielewicz_2015}.
Post-mortem iris images were selected from \emph{Warsaw-BioBase-Post-Mortem-Iris v1.0} dataset \cite{Trokielewicz_2016}.
Iris~images of twins and images presenting high difference in pupil dilation were selected from datasets of the University of Notre Dame, including the one used by Hollingsworth~\etal~\cite{hollingsworth_2011}.

\subsection{Notre Dame Experiments}
\label{sec:nd}

Custom software was prepared for both annotation-less and annotation-driven sessions.
In the annotation-less \emph{Session~1}, 86 adult individuals (between 18 and 65 years old) from the university community (including students, staff, and faculty) volunteered to participate, with no constraints related to gender and ethnicity.
All were subject to the same protocol, approved by the internal academic \emph{Human Subjects Institutional Review Board}.
Each volunteer was asked to evaluate a set of 20 iris image pairs, which always contained the following distribution of image pairs, presented in randomized order for each subject:

\begin{itemize}
    \item four healthy easy pairs, with two impostor and two genuine samples;
    \vspace{-0.2cm}
    
    \item four healthy difficult pairs, with two impostor and two genuine samples;
    \vspace{-0.2cm}
    
    \item four genuine pairs of irises with large difference in pupil dilation;
    \vspace{-0.2cm}
    
    \item four impostor twins' pairs;
    \vspace{-0.2cm}
    
    \item four genuine post-mortem pairs.
    \vspace{-0.2cm}
\end{itemize}

In each trial, the software displayed a pair of iris images and asked the subject to select one of the following: \emph{``1.~same person (certain)''}, \emph{``2.~same person (likely)''}, \emph{``3.~uncertain''}, \emph{``4.~different people (likely)''}, and \emph{``5.~different people (certain)''}.

\begin{figure}[t]
\centering
\frame{\includegraphics[width=0.47\textwidth]{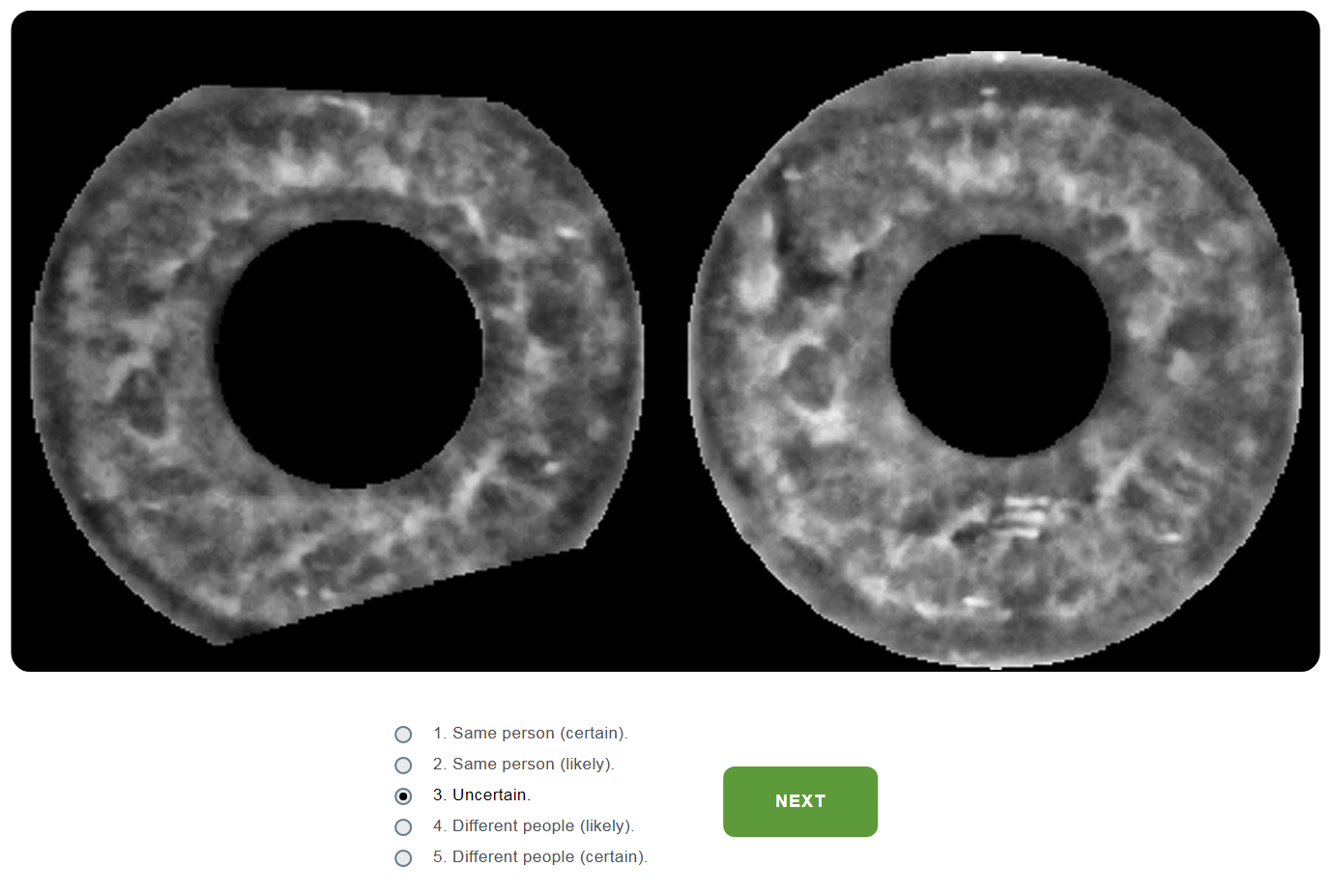}}
\caption{An example screen of a \emph{Session~1} trial. To proceed to the next image pair, the subject had to select one decision and click ``Next''. }
\label{fig:session-1-printscreen}
\vspace{-0.1cm}
\end{figure}

Fig.~\ref{fig:session-1-printscreen} depicts the interface of the software, showing one of the 20 trials and possible answers.
Each subject could spend as much time as necessary before selecting the answer.
The following trial was only displayed after the acceptance of the current selection.
In total, the software used 20 disjoint sets of 20 image pairs, leading to 400 available trials composed from 800 manually segmented iris images.
As a consequence, each trial set was submitted, on average, to four subjects, but never with the same order, since, for each individual, the tool randomly shuffled the 20 trials to be presented.
On average, each subject spent seven minutes participating in the annotation-less first session.

\begin{figure}[t]
\centering
\frame{\includegraphics[width=0.47\textwidth]{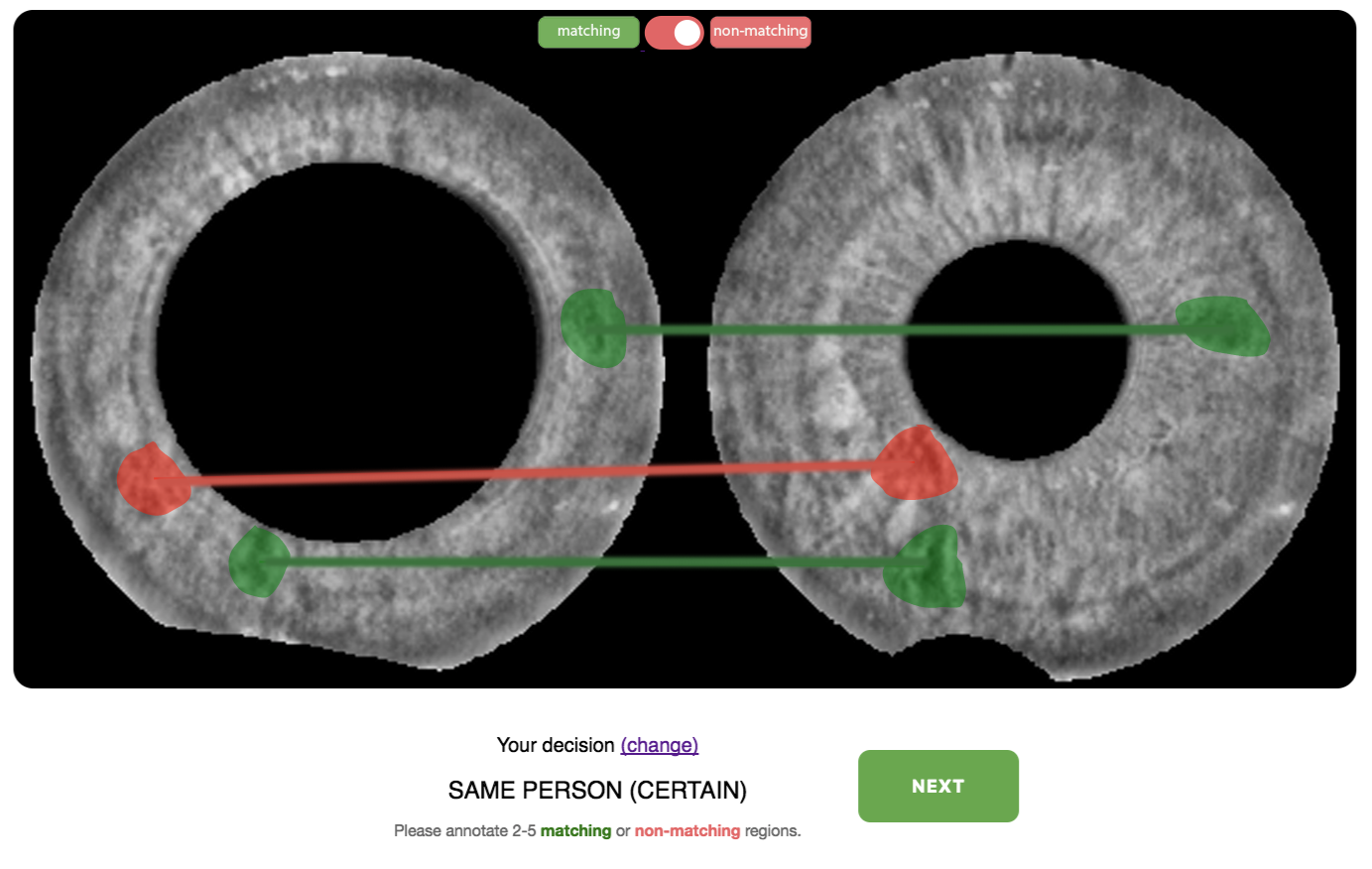}}
\caption{An example screen of a \emph{Session~2} trial.
Subjects were allowed to freely annotate and connect matching (green) or non-matching (red) features over the two presented irises, to support their decision.
They could update their previous decision by clicking on the ``change'' option.}
\label{fig:session-2-printscreen}
\vspace{-0.4cm}
\end{figure}

In the second session, 85 of the 86 subjects continued to provide manual annotations for both matching and non-matching regions between irises.
For each person, from the previous 20 trials they have opined, the software automatically selected 10 trials in an iris-category balanced manner.
In addition, the tool tried to present, if possible, at least one hit and one miss of each category.
Subjects were allowed to annotate as many pairs of matching or non-matching regions as they want, however annotating from 2 to 5 feature pairs was recommended. Also, they were advised to avoid using masked out black regions.
Fig.~\ref{fig:session-2-printscreen} depicts the annotation interface of the custom software, showing an example annotation.
Subjects could freely change and update their decisions while annotating a particular pair.
Each subject spent between 10 and 20 minutes participating in the annotation-driven session.

\subsection{NASK Experiments}
\label{sec:wu}
\vspace{-0.1cm}

These experiments consisted of only the annotation-less (first) session.
In total, 28 subjects (different from those attending the experiments at Notre Dame) participated, committing themselves to the exact same protocol, locally approved by the NASK data-protection office.
Each subject was asked to evaluate a set of 24 image pairs, which always contained the following setup:
\vspace{-0.1cm}

\begin{itemize}
    \item five genuine post-mortem iris pairs;
    \vspace{-0.16cm}
    
    \item five impostor post-mortem iris pairs;
    \vspace{-0.16cm}
    
    \item five genuine disease-affected iris pairs;
    \vspace{-0.16cm}
    
    \item five impostor disease-affected iris pairs;
    \vspace{-0.16cm}
    
    \item four repeated pairs, each one being randomly selected from one of the above subsets.
    \vspace{-0.16cm}
\end{itemize}

In each trial, a custom software displayed a pair of iris images and asked the subject to provide a binary decision on whether images depicted the same eye or not.
For the 28 subjects, the software had 10 disjoint sets of 24 trials available, leading to a total of 240 available iris pairs.
As a consequence, each trial set was presented to at least two subjects.
On average, each subject spent 15 minutes participating in this experiment.

\subsection{OSIRIS, IriCore, and MIRLIN Setup}
\label{sec:os}
\vspace{-0.1cm}

We used OSIRIS~\cite{othman_2016}, IriCore~\cite{iricore_2018}, and MIRLIN~\cite{mirlin_2018} as representatives of automated iris-matching algorithms.
OSIRIS implements a Daugman-style solution~\cite{daugman_2004}, therefore relying on Gabor filters to generate iris codes that are compared through fractional Hamming distance.
IriCore and MIRLIN, in turn, comprise two commercial iris recognition solutions, which together represent the current state of the art in this area.
All three methods generate genuine comparison scores that should be close to zero.
Since OSIRIS does not apply Daugman's score normalization~\cite{daugman_2006}, we assumed an acceptance threshold equal to 0.32, as earlier suggested in~\cite{daugman_2004}.
With respect to IriCore, we adopted an acceptance threshold of 1.1, as suggested in its documentation.
For MIRLIN, in turn, we used a threshold of 0.2, as recommended in~\cite{czajka_2016}.

\section{Results}
\label{sec:results}
\vspace{-0.1cm}

Table~\ref{tab:session-1-acc} shows the performance of human subjects in assessing the comparison type (genuine or impostor) of iris pairs during the first annotation-less session of experiments, combined for both \emph{Notre Dame} and \emph{NASK} experiments.
Reported accuracy expresses the percentage of correctly classified trials, across all subjects.
A subject's response was considered correct, or a \emph{hit}, if the subject selected ``same person'', with either ``certain'' or ``likely'' as their confidence, and the image pair was in fact from the same iris.
A ``different people'' response, with either ``certain'' or ``likely'' confidence, was considered correct, or also a \emph{hit}, if the image pair was in fact of different irises.
All other responses, including the ``uncertain'' option, were treated as a mistake, \ie a \emph{miss}.
Given that people's decisions were discrete, we do not provide ROC-like curves in the results.
In addition, OSIRIS, IriCore, and MIRLIN were used according to their respective recommended operating points.

\begin{table}[t]
\caption{
Annotation-less performance of human subjects in iris identification.
Subjects were only asked to select their decisions.
For comparison sake, we report results of the OSIRIS, IriCore, and MIRLIN software, with acceptance thresholds equal to 0.32, 1.1, and 0.2, respectively.
}
\label{tab:session-1-acc}
\centering
\footnotesize
\begin{tabular}{C{0.26cm} R{1.9cm} C{0.94cm} C{0.9cm} C{0.82cm} C{1.0cm}} 
\hline
& \multirow{2}{*}{Iris category}  & \multicolumn{4}{c}{Accuracy (\%)}\\
& & Humans & OSIRIS & IriCore & MIRLIN \\
\hline
\multirow{6}{*}{\rotatebox[origin=c]{90}{Genuine pairs}} & Healthy easy & 91.28 & 95.00 & 100.00 & 97.50 \\
 & Healthy difficult & 79.07 & 90.00 & 97.50 & 97.50 \\
 & Pupil-dynamic & 43.90 & 61.25 & 95.00 & 97.50 \\
 & Post-mortem & 51.95 & 33.57 & 73.57 & 47.14 \\ 
 & Disease-affected & 70.80 & 25.00 & 53.33 & 25.00 \\
\cmidrule(lr){2-6}
 & Combined & \textbf{60.60} & \textbf{58.86} & \textbf{80.56} & \textbf{65.83}\\
\hline
\multirow{6}{*}{\rotatebox[origin=c]{90}{Impostor pairs}} & Healthy easy & 84.30 & 100.00 & 100.00 & 100.00 \\
 & Healthy difficult & 76.16 & 100.00 & 100.00 & 100.00 \\
 & Twins & 55.81 & 100.00 & 100.00 & 100.00 \\
 & Post-mortem & 83.90 & 100.00 & 100.00 & 100.00 \\
 & Disease-affected & 91.00 & 100.00 & 100.00 & 100.00 \\
\cmidrule(lr){2-6}
 & Combined & \textbf{74.41} & \textbf{100.00} & \textbf{100.00} & \textbf{100.00} \\
\hline
\multicolumn{2}{l}{Overall} & \textbf{70.11} & \textbf{79.43} & \textbf{89.06} & \textbf{80.78}\\
\hline
\end{tabular}
\end{table}

Overall, subjects were correct nearly 70\% of the time, while the best algorithm (IriCore) achieved a higher overall accuracy of 89.06\%.
Both human subjects and software tools were more successful in identifying impostor than genuine pairs, with all the three tools not making a single mistake in recognizing impostors.
Nonetheless, in the particular case of genuine samples, humans performed on par with OSIRIS and MIRLIN, exceeding their results in face of genuine post-mortem samples.
Moreover, people presented a much superior performance when analyzing disease-affected samples, surpassing all the three tools.
Indeed, in such cases, software was always biased towards classifying samples as impostors, justifying close-to-chance (IriCore) or worse-than-chance (OSIRIS, MIRLIN) accuracies for genuine pairs, and perfect hit rates for impostors.

Subjects performed better than chance, \ie the accuracy was higher than 50\%, in most of the iris categories.
However, for the subset composed of iris images with large difference in pupil dilation, the accuracy was only 43.9\%.
Variations in pupil dilation were the most challenging cases for subjects, impairing their ability to recognize different versions of the same eye.

Subjects also had difficulty in analyzing post-mortem iris pairs.
In general, they tended to classify post-mortem samples as impostors, leading to a low accuracy in genuine cases (51.95\%, slightly better than chance), and a higher accuracy in impostor cases (83.90\%).
Similar to the observations of Bowyer \etal~\cite{bowyer_2010}, irises of twins also revealed themselves as challenging for people, but easy for automated solutions.
Among impostor samples, they are the category where subjects had the lowest accuracy (55.81\%).

\begin{figure}[t]
\centering
\includegraphics[width=8.3cm]{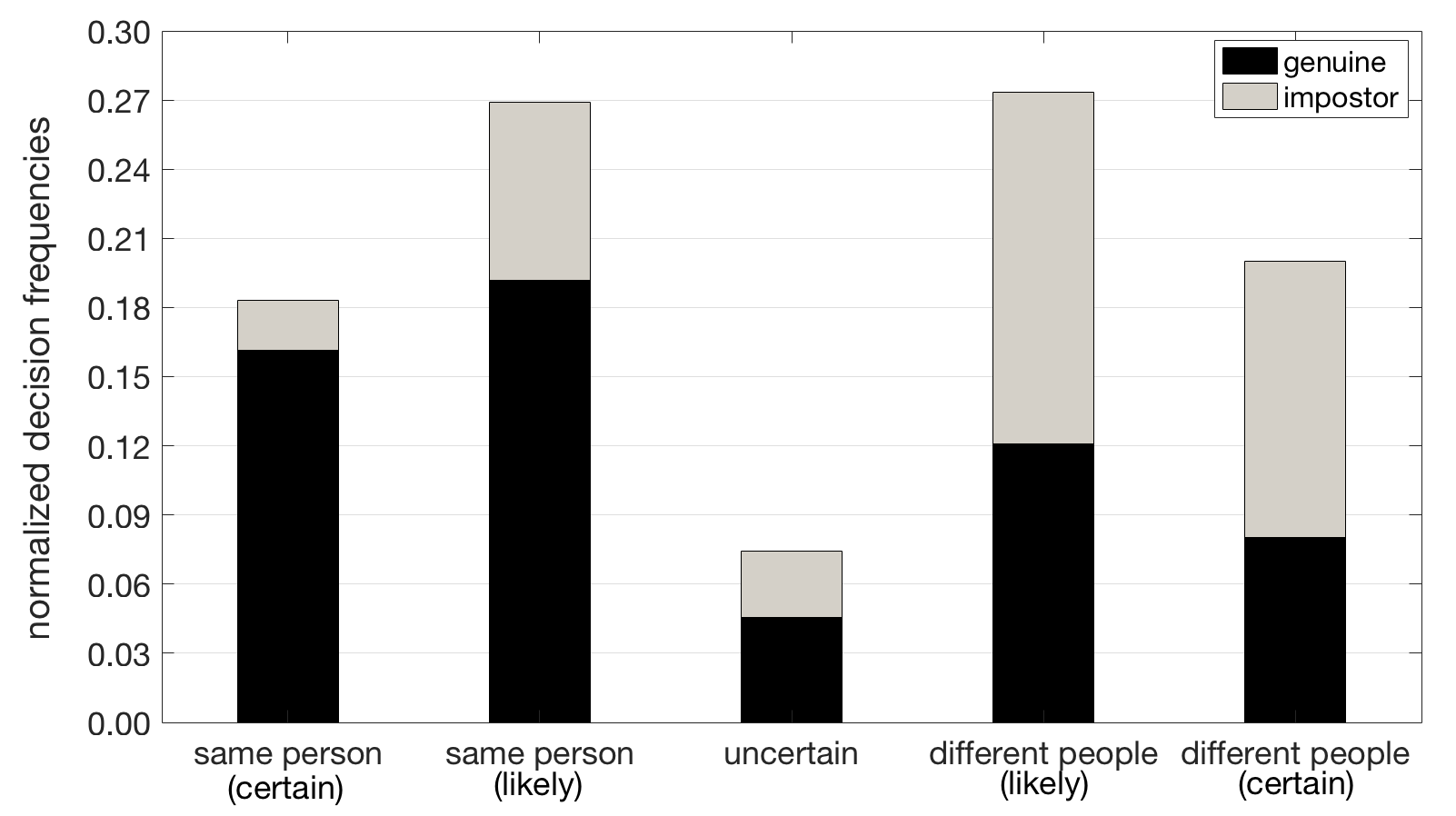}
\caption[]{
Normalized frequencies of the decisions of the 86 subjects of the \emph{Notre Dame} experiments, according to their decisions and groundtruth.
Genuine iris pairs are represented by black bars, while impostor pairs are represented by gray ones.
In an ideal classification output, black bars should happen only on the left part of the chart, while gray bars should happen only on the right side.
}
\label{fig:opinion-conviction}
\vspace{-0.4cm}
\end{figure}

Fig.~\ref{fig:opinion-conviction} shows the subjects' confidence level when classifying the iris pairs, during the first session of \emph{Notre Dame} experiments.
Bars depict the normalized frequencies of each response; as a consequence, they sum up to 1.0.
According to the adopted groundtruth color notation, black-bar regions represent genuine pairs and gray regions represent impostor pairs.
Therefore, black regions are expected to occur mostly on the left side of the chart (which is respective to people's claims of seeing genuine pairs), while gray regions are expected to occur mostly on the right side (which is respective to impostor pairs).
Gray regions on the left side and black regions on the right side are all errors, as well as any answer in the center (``uncertain'' option).

As one might observe, the ``uncertain'' option was the least selected choice (being taken in less than 8\% of the trials), agreeing with the reports of McGinn et al.~\cite{mcginn_2013}.
Among the ``same person (certain)'' answers (18\% of all answers), one in each nine (nearly 11\%) was wrong.
In opposition, among the ``different people (certain)'' answers (20\% of all answers), nearly one third was wrong, revealing more errors in people's convictions of seeing impostor pairs.
In accordance to the data presented in Table~\ref{tab:session-1-acc}, this indicates that people had more problems in recognizing genuine pairs, wrongly classifying many of them as impostors with high confidence.

Table~\ref{tab:session-2-acc} provides a comparison of the performances of 85 subjects who participated in the first session of \emph{Notre Dame} experiments (without annotations), and were able to return to the second session, when they were asked to provide annotations for the matching and non-matching regions between the irises of each presented pair.
Reported accuracy regards the correctness of the decisions for the subset of iris pairs they have already seen in the first session.

\begin{table}[t]
\caption{
Comparison of 85 subjects' accuracy when performing iris identification without annotations versus with annotations, over exactly the same iris samples.
}
\label{tab:session-2-acc}
\centering
\footnotesize
\begin{tabular}{C{1.1cm} R{1.9cm} C{1.5cm} C{1.5cm}}
\hline
\multirow{3}{*}{Pair class} & \multirow{3}{*}{Iris category} & \multicolumn{2}{c}{Accuracy (\%)}\\
 & & without annotations & with annotations\\
\hline
\multirow{5}{*}{Genuine} & Healthy easy & 87.06 &96.47\\
 & Healthy difficult & 75.29 & 84.71\\
 & Pupil-dynamic & 41.18 & 52.35\\
 & Post-mortem & 45.29 & 54.12\\
\cmidrule(lr){2-4}
 & Combined & \textbf{55.88} & \textbf{65.69}\\
\hline
\multirow{4}{*}{Impostor} & Healthy easy & 85.88 & 90.59\\
 & Healthy difficult & 80.00 & 78.82\\ 
 & Twins & 59.41 & 60.59\\
\cmidrule(lr){2-4}
 & Combined & \textbf{71.18} & \textbf{72.65}\\
\hline
 Overall & & \textbf{62.00} &\textbf{68.47}\\
\hline
\end{tabular}\\
\vspace{-0.2cm}
\end{table}

The annotation feature helped subjects to improve their decisions in all iris categories, except for impostor healthy difficult pairs (in which accuracy slightly dropped from 80.00\% to 78.88\%).
Iris image pairs with large difference in pupil dilation were the category that benefited the most from annotations, with an improvement in accuracy from 41.18\% to 52.35\%.
Accuracy for post-mortem cases also was significantly improved.

\begin{figure}[t]
\centering
\includegraphics[width=8.3cm]{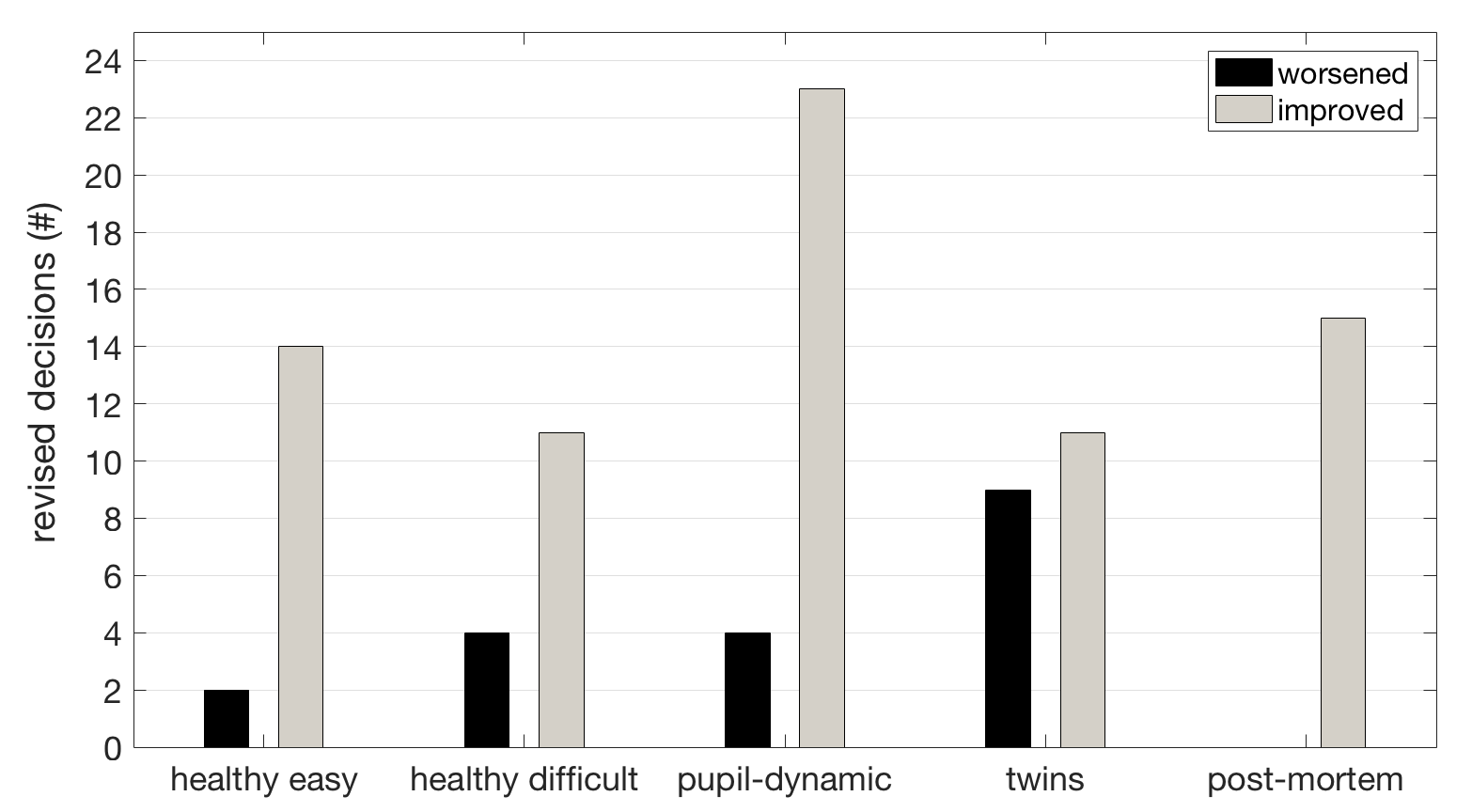}
\caption[]{
Numbers of revised iris pairs grouped by iris category.
While manually annotating an iris pair, subjects could change their decision, either improving it (\ie making it right, depicted in gray), or worsening it (\ie making it wrong, depicted in black).
}
\label{fig:revised-opinions}
\vspace{-0.2cm}
\end{figure}

Fig.~\ref{fig:revised-opinions} details how decisions were revised when subjects provided manual annotation.
Black bars express the absolute number of revised decisions that were worsened (\ie, a correct decision after the first session, updated to an incorrect decision during the second session).
Conversely, gray bars express the number of revised decisions that were fixed (\ie, they were originally a miss after first session, but then were updated to a correct decision during the second one).
In general, more decisions were fixed (74 decisions) than worsened (19 decisions).
Interestingly, post-mortem samples presented only improvements (15 decisions were revised), suggesting that people perceived new details on them while providing annotations.
Twins' samples, in turn, once more revealed how confusing they appear to people; 11 incorrect decisions were corrected, but 9 correct decisions were changed to be incorrect.

\begin{figure}[t]
\centering
\includegraphics[width=8.3cm]{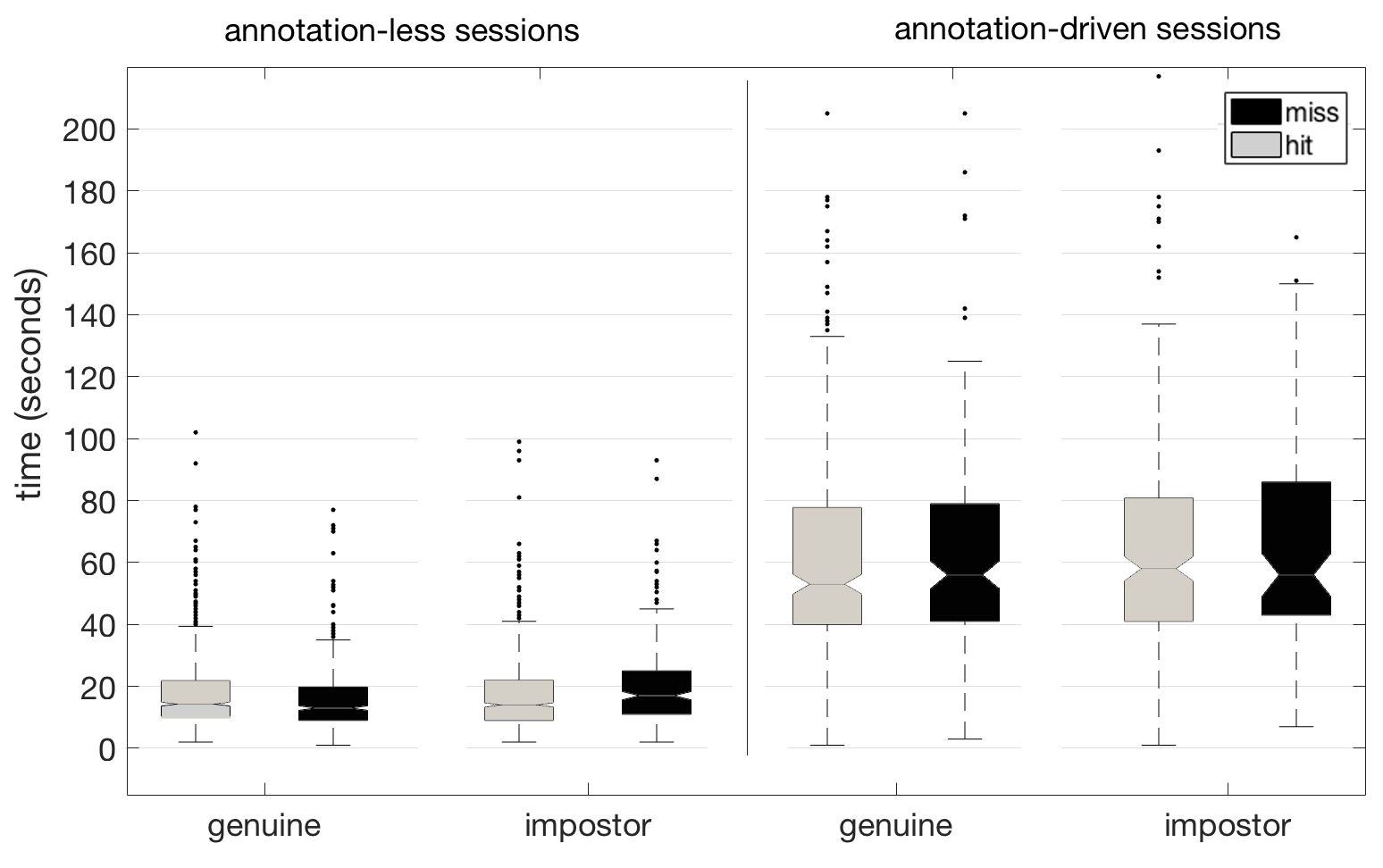}
\caption[]{
Times spent by subjects to answer each trial.
Left side: times spent in the annotation-less sessions.
Right side: times spent in the annotation-driven sessions.}
\label{fig:time-acc}
\vspace{-0.2cm}
\end{figure}

Last but not least, we could not find correlation between time spent by subjects and accuracy.
Fig.~\ref{fig:time-acc} depicts the distributions of time spent by subjects to decide each trial in the first annotation-less sessions (combining both \emph{Notre Dame} and \emph{NASK} experiments, shown in the left side of the chart), and to annotate each trial in the second annotation-driven sessions (shown in the right side of the chart).
As one might observe, regardless of iris pairs being genuine or impostor, and of decisions being hits or misses, distributions were not significantly different.
As expected, annotation-driven trials were, on average, longer than annotation-less trials.

\section{Conclusions}
\label{sec:conclusions}
\vspace{-0.1cm}

This paper presents results of a unique study estimating the accuracy of human subjects in comparing iris images of different levels of difficulty, including healthy and disease-affected eyes, and images acquired from cadavers.

The first observation from this study is that we may expect people to be worse than automated iris-recognition methods when comparing healthy eyes.
However, they can be better in cases not yet considered in the development of automated algorithms, such as eyes suffering from diseases or post-mortem deformations. 

The second observation is that human examiners on average improve their accuracy when they are asked to annotate matching and non-matching features that support their decision.
Although this improvement is larger for genuine pairs than for impostor pairs, it still suggests that a comparison of iris images performed by humans should be organized in a way that allows them to annotate the features they are using in their judgment.
As future work, this may help in the development of a method for the examination and documentation of irises that is analogous to \emph{ACE-V}~\cite{ashbaugh_1999}, originally proposed for fingerprints.

The third observation is that different categories of iris images result in significantly different performance of human subjects.
Three categories of samples that seem to be particularly challenging to humans: irises of identical twins, iris images showing large difference in pupil dilation, and irises of deceased individuals.

{\small
\bibliographystyle{ieee}
\bibliography{ref}
}

\end{document}